\documentclass{article} 
\usepackage{iclr2022_conference,times}

\usepackage{amsmath,amsfonts,bm}

\def\eqref#1{equation~\ref{#1}}

\def\1{\bm{1}}

\def\vtheta{{\bm{\theta}}}

\def\vx{{\bm{x}}}

\DeclareMathAlphabet{\mathsfit}{\encodingdefault}{\sfdefault}{m}{sl}
\SetMathAlphabet{\mathsfit}{bold}{\encodingdefault}{\sfdefault}{bx}{n}

\usepackage{hyperref}
\usepackage{url}
\usepackage{times}
\usepackage{epsfig}
\usepackage{graphicx}
\usepackage{amsmath}
\usepackage{amssymb}
\usepackage{array}
\usepackage{multirow}
\usepackage{booktabs}
\usepackage{caption}
\usepackage{subcaption}
\usepackage{color,soul}

\title{Dynamically Decoding Source Domain Knowledge for Domain Generalization}

\author{Cuicui Kang and Karthik Nandakumar \\
Department of Computer Vision\\
Mohamed bin Zayed University of Artificial Intelligence\\
Masdar City, Abu Dhabi, UAE \\
\texttt{\{Cuicui.Kang, Karthik.Nandakumar\}@mbzuai.ac.ae} \\
}

\iclrfinalcopy 
\begin{document}

\maketitle

\begin{abstract}
Optimizing the performance of classifiers on samples from unseen domains remains a challenging problem. While most existing studies on domain generalization focus on learning domain-invariant feature representations, multi-expert frameworks have been proposed as a possible solution and have demonstrated promising performance. However, current multi-expert learning frameworks fail to fully exploit source domain knowledge during inference, resulting in sub-optimal performance. In this work, we propose to adapt Transformers for the purpose of dynamically decoding source domain knowledge for domain generalization. Specifically, we build one domain-specific local expert per source domain and one domain-agnostic feature branch as query. A Transformer encoder encodes all domain-specific features as source domain knowledge in memory. In the Transformer decoder, the domain-agnostic query interacts with the memory in the cross-attention module, and domains that are similar to the input will contribute more to the attention output. Thus, source domain knowledge gets dynamically decoded for inference of the current input from unseen domain. This mechanism enables the proposed method to generalize well to unseen domains. The proposed method has been evaluated on three benchmarks in the domain generalization field and shown to have the best performance compared to state-of-the-art methods.
\end{abstract}

\section{Introduction}
\label{sec:introduction}

Domain shift is a common phenomenon in most real-world applications \citep{WangDGSurvey2021,ZhouDGSurveyVision2021}. Hence, it is critical to develop computer vision systems that generalize well across domains not observed during training. While many strategies such as transfer learning and domain adaptation have been proposed to address domain shift, domain generalization has received significant attention in the recent past \citep{LiArtierDG2017}. The goal in domain generalization (DG) is to learn models based on one or more source domains, which can perform accurately on unseen target domains.

Algorithms for DG can be broadly grouped into three categories \citep{WangDGSurvey2021}. While the first set of methods attempt to learn robust feature representations that work well across domains \citep{MuandetInvariantFeatures2013,GhifaryMTAE2015,GaninDomainAdversarialTraining2016}, the second approach relies on data augmentation to enhance the diversity of the training data based on images available in the given source domains \citep{ShankarCrossGrad2018,VolpiADA2018,ZhouMixStyle2021}. The third category focuses on the learning strategy, which includes meta-learning \citep{FinnModelAgnosticMetaLearning2017,LiMLDG2018,BalajiMetaReg2018} and learning a mixture of experts \citep{DInnocenteDSAM2018,ManciniBestSourcesForward2018,DaiRAMoE2021,ZhouDAEL2021}. A combination of data augmentation and other innovations in training have been proposed in \citep{CarlucciJiGen2019,LiAGGEpiFCR2019,HuangSelfChallenging2020}. This work follows the mixture of experts learning strategy, which is based on the intuition that a test sample from an unseen domain may still share some similarity with the available source domains. Therefore, if the domain relationships can be learned, closely related domain expert features should be able to infer the test image well.

\begin{figure*}[ht]
\centering
\begin{subfigure}[b]{0.4\textwidth}
\includegraphics[width=0.96\textwidth]{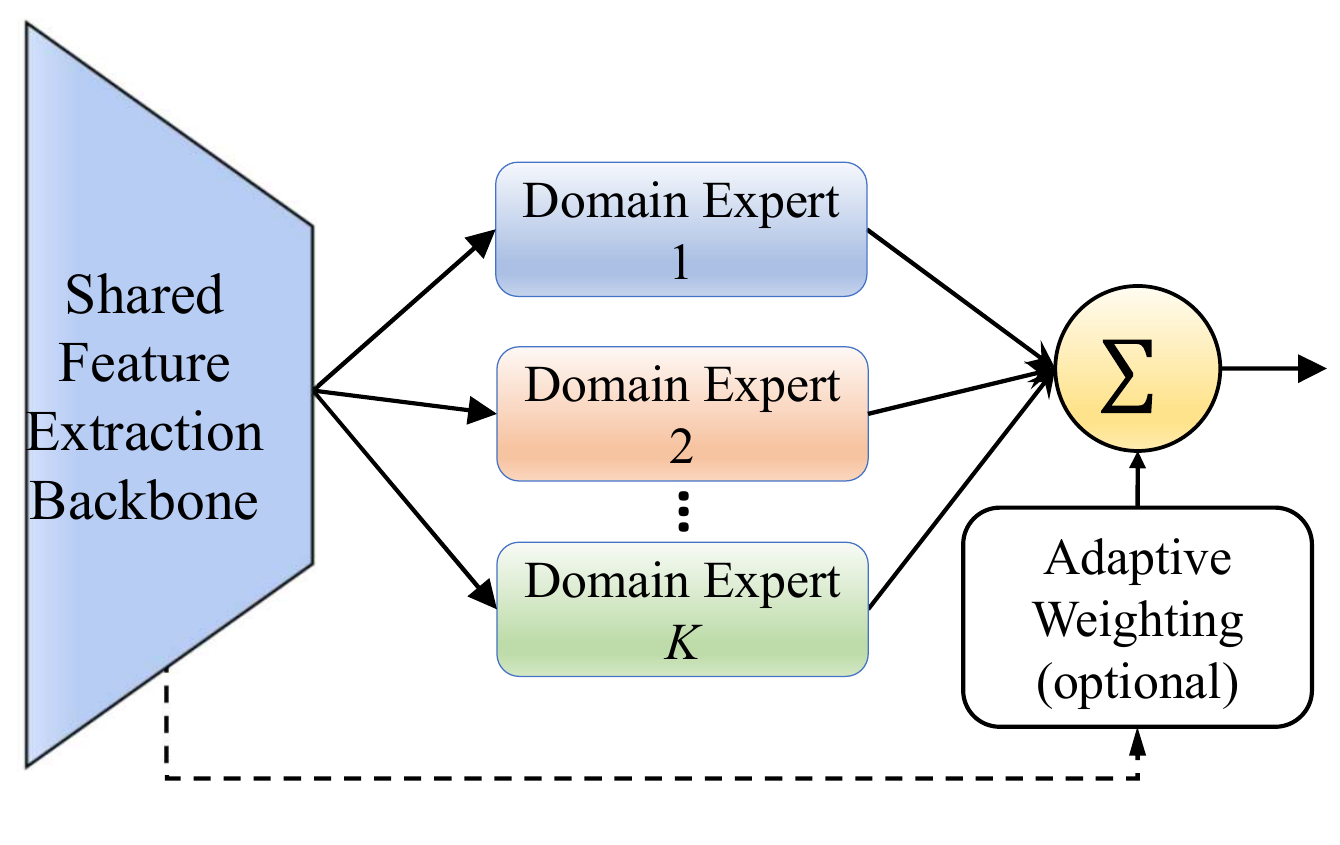}
\caption{}\label{fig:ConventionalMoE}
\end{subfigure}
\hspace{0.2in}
\begin{subfigure}[b]{0.53\textwidth}
\includegraphics[width=0.98\textwidth]{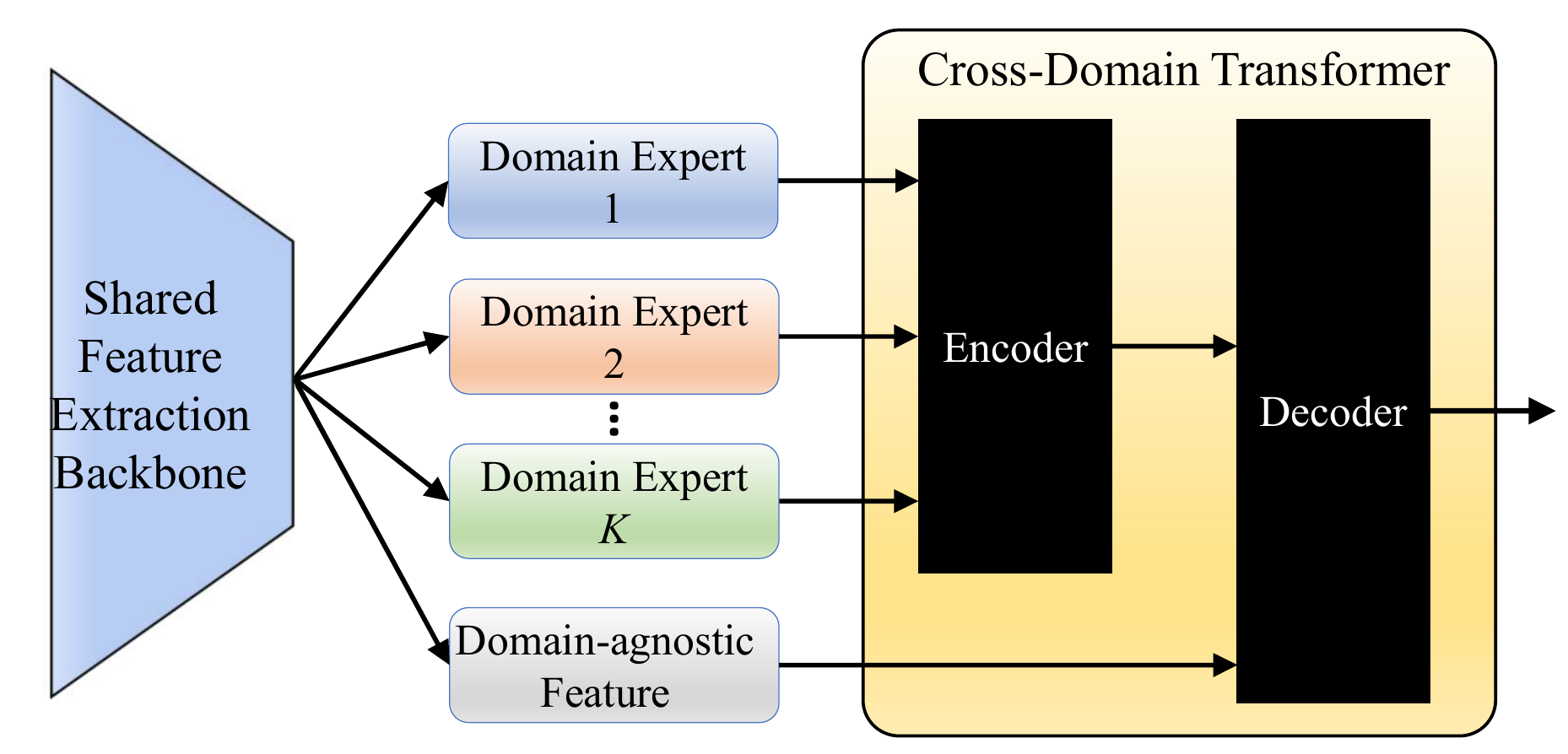}
\caption{}\label{fig:simple}
\end{subfigure}
\caption{Mixture of Experts (MoE) for domain generalization. While existing methods (a) use a weighted combination of domain experts, the proposed method (b) uses a cross-domain Transformer to encode source domain knowledge and dynamically decode it during inference.}
\end{figure*}

Existing mixture of experts (MoE) methods \citep{ManciniBestSourcesForward2018,DInnocenteDSAM2018,ZhouDAEL2021,DaiRAMoE2021} for DG perform inference mostly based on a weighted linear combination of the domain experts, where the weights may be dynamically assigned (see Figure \ref{fig:ConventionalMoE}). The key differences between these methods lie in how the domain-specific local experts are learned and how they are aggregated to represent the target unseen domain image. In the existing MoE algorithms, the linear nature of the combination function is a serious limitation. Consider the case of an unseen domain that is a merger of two source domains (e.g., part photo and part art-painting). In this case, the weighting approach will determine that both the source domains are equally relevant and average out the domain-specific features from both domains. However, a more appropriate strategy is to identify a subset of relevant features from each of the two domains and combine them. Furthermore, existing MoE methods lack knowledge transfer or interactions between the different domains during inference. Consequently, the correlations and complementarities between the feature representations of individual domains (conditional on the given input sample from target domain) are largely ignored.

To address the above limitations, we propose a hybrid deep architecture of domain-specific local experts and Transformer-based query-memory decoding for domain generalization, as shown in Figure \ref{fig:simple}. Together with domain-specific local experts and a domain-agnostic query feature branch, a cross-domain Transformer is designed to dynamically decode source domain knowledge for inference of a new image from the unseen domain. In the proposed architecture, a shared Convolutional Neural Network (CNN) backbone is utilized for efficiently learning generic low-level features. A domain-specific expert for each source domain as well as a domain-agnostic query feature branch are learned based on the shared features. The cross-domain Transformer facilitates deep feature learning and exploration of domain relationships. The source domain features are encoded as memory, where self-attention enables interaction among different domains. The domain-agnostic feature branch is used as query, and a Transformer decoder is applied with both memory (keys and values) and query as inputs to identify the relevant knowledge available in the source domains and dynamically decode the source domain knowledge for inference of the input image from unseen domain. The final feature representation output by the cross-domain Transformer is used for classification.

The contributions of this work can be summarized as:

\begin{itemize}
    \item Design of an architecture that encodes source domain knowledge by applying Transformer encoder to outputs of domain-specific local experts. 

    \item A Transformer decoding scheme that allows interaction between the domain-agnostic query and the encoded memory in the cross-attention module, where domains that are similar to the input will contribute more to the attention output. Thus, the source domain knowledge gets dynamically decoded for inference of new images from unseen domains, which makes the proposed method generalize well to target domains.

    \item Evaluation of the proposed method in the context of DG for object classification, which demonstrates the effectiveness of the proposed approach.
\end{itemize}

\section{Related Work}
\label{sec:relatedwork}

\noindent \textbf{Domain Generalization}: Comprehensive surveys on domain generalization have been published recently in \citep{WangDGSurvey2021,ZhouDGSurveyVision2021}. The traditional approach to DG involves learning domain-invariant representations. For instance, \cite{MuandetInvariantFeatures2013} tried to learn an invariant transformation for DG by minimizing the differences in the marginal distributions across source domains and a kernel-based method was proposed. Multi-Task Auto-Encoder (MTAE) was used to learn unbiased object features by \cite{GhifaryMTAE2015}. In order to minimize the distance between images from the same category but different domains, maximum mean discrepancy was used in \citep{MotiianDeepSDA2017} to align feature distributions. Recently, adversarial training has become a popular strategy for learning domain-invariant features \citep{GaninDomainAdversarialTraining2016,LiMMD2018,ZhaoER2020,MatsuuraMMLD2020}. Another variant of this approach is the learning of domain-invariant model parameters \citep{KhoslaUndoingDatasetBias2012,LiArtierDG2017}.

Data augmentation is another well-known approach to improve the domain generalization ability. A BayesianNet was used by \cite{ShankarCrossGrad2018} to perturb the input data and generate more diverse data samples for training. ``Hard'' training samples were synthesized in \citep{VolpiADA2018} and samples with new styles were created in \citep{ZhouMixStyle2021} by mixing up the styles of training instances. Multiple source domains and categories were mixed up in \citep{ManciniCuMix2020} to produce unseen categories in unseen domain to solve both DG and zero-shot learning. Optimal transport was for data synthesis in \citep{ZhouLTAOT2020}. The data augmentation approach can also be combined with other innovative learning strategies such as self-supervised learning \citep{CarlucciJiGen2019}, episodic training \citep{LiAGGEpiFCR2019}, and self-challenging \citep{HuangSelfChallenging2020} to achieve domain generalization.

Inspired by the success of the Learning to Learn or meta learning paradigm, the MLDG technique was proposed in \citep{LiMLDG2018} for DG. The main idea underlying these methods is to divide the given source domains into meta-train and meta-test subsets and simulate the domain shift problem during training, which can be solved using the various optimization strategies \citep{SantoroMetaLearningMANN2016,FinnModelAgnosticMetaLearning2017}. A regularization function based on meta learning called MetaReg was proposed in \cite{BalajiMetaReg2018} to improve DG and model-agnostic learning of semantic features (MASF) was introduced in \citep{DouMASF2019}. Recently, it has been argued that standard empirical risk minimization (ERM) approach based on baseline CNNs can provide comparable performance to state-of-the-art DG algorithms, provided the model selection is done fairly \citep{GulrajaniISLDG2021}. Moreover, ERM with dense stochastic weight averaging (SWAD) to identify flat minima has also been shown to be quite effective \citep{ChaSWAD2021}.

\noindent \textbf{Mixture of Experts for Domain Generalization}: Suppose that there are $K$ source domains. Let $f$ represent a shared (domain-agnostic) feature extraction module and $g_k$ denote the domain-specific expert for source domain $k, k=1,2,\cdots,K$. MoE algorithms typically obtain the output for a given test sample $\vx$ as a function of $\sum_{k=1}^{K}w_k(f(\vx))g_k(f(\vx))$. In the BSF algorithm \citep{ManciniBestSourcesForward2018}, there is no shared feature extraction, i.e., $f(\vx)=\vx$, and the output is a function of $\sum_{k=1}^{K}w_k(\vx)g_k(\vx)$, where $g_k$ is the domain-specific classifier and $w_k$ measures the relevance of the source domain to the unseen test domain. In contrast, the D-SAM \citep{DInnocenteDSAM2018}, DSON \citep{SeoDSN2020}, and DAEL \citep{ZhouDAEL2021} algorithms ignore the weights and output a function of $\sum_{k=1}^{K}g_k(f(\vx))$. While the D-SAM method learns domain-specific aggregation modules $g_k$ based on a pre-trained feature extractor $f$, the DSON technique performs domain-specific optimized normalization $g_k$, and the DAEL model learns the domain-specific classification heads $g_k$ in a collaborative fashion, with each local expert providing a soft supervision signal to the ensemble of non-experts learned from the other source domains. Finally, in relevance-aware mixture of experts (RaMoE) \citep{DaiRAMoE2021}, a separate voting network is used to learn ``query'' features from the test image and the weights $w_k(f(\vx))$ are determined based on the domain relevance score (computed as the inner product of the query features and domain-specific features). All the above methods are limited in their ability to robustly characterize complex domain relationships, , especially during inference.

\noindent \textbf{Transformers}: Transformer is a global attention method that was first proposed in \citep{VaswaniTransformers2017} for machine translation. The core module in the Transformer architecture is Multi-Head Attention (MHA), which models the relationships between a sequence of symbol representations based on scaled dot-product attention. The Transformer architecture consists of two main components - an encoder and a decoder, each composed of a stack of $L$ identical layers. Each encoder layer has two blocks - a Multi-Head Self Attention (MHSA) block and a Multi Layer Perceptron (MLP) block. In the MHSA block of an encoder, the queries, keys, and values are identical. For each block, both layer normalization and residual connections are added to stabilize the learning of the deep architecture. In contrast, each decoder layer has three blocks - two MHA blocks and a MLP block. While the first MHA block in the decoder performs self-attention over the output embeddings, it is followed by a Multi-Head Cross Attention (MHCA) block, where the keys and values come from the output of the encoder (considered as memory) and are different from the query. Similar to the encoder, each decoder block also employs residual connections, followed by layer normalization. Since the Transformer architecture is permutation-invariant, positional encodings are concatenated to each symbol representation in the sequence.

Inspired by the success of Transformers in natural language processing, Vision Transformers (ViT) \citep{DosovitskiyViT2021} were introduced for image classification by applying Transformer encoder to raw image patches. Bottleneck Transformers that replace some of the spatial convolutions in a deep CNN with MHSA blocks were proposed in \citep{SrinivasBottleneckTransformer2021}. Both these methods utilize only the self-attention mechanism. The Detection Transformer (DETR) \citep{CarionDETR2020} uses both Transformer encoder and decoder for object detection, but the decoder uses learned position embeddings as object queries. The proposed method employs both self attention and cross attention mechanisms. The decoder takes an initial representation of the test sample as the query to find the relevant discriminative features based on knowledge extracted by the encoder from the available source domains.

\section{Proposed Method}
\label{sec:proposedmethod}

\subsection{Problem Definition}
In multi-source domain generalization, the goal is to learn a model using data from $K$ source domains $\mathcal{D}_S = \{\mathcal{D}_k\}_{k=1}^{K}$ that generalizes well to data from an unseen target domain $\mathcal{D}_T$. It is assumed that all the domains share the same input space $\mathcal{X}$ and label space $\mathcal{Y}$, but have different data distributions. Since this work deals with object recognition, $\mathcal{Y} = \{1,2,\cdots,N_C\}$, where $N_C$ is the number of object categories. The training data can be represented as $\mathcal{D}_S = \{\vx_i,y_i,z_i\}_{i=1}^{N_S}$, where $\vx_i \in \mathcal{X}$, $y_i \in \mathcal{Y}$, $z_i$ is the domain to which sample $i$ belongs, and $N_S$ is the cardinality of the training set.

\begin{figure*}[t]
\centering
    \includegraphics[width=0.98\linewidth]{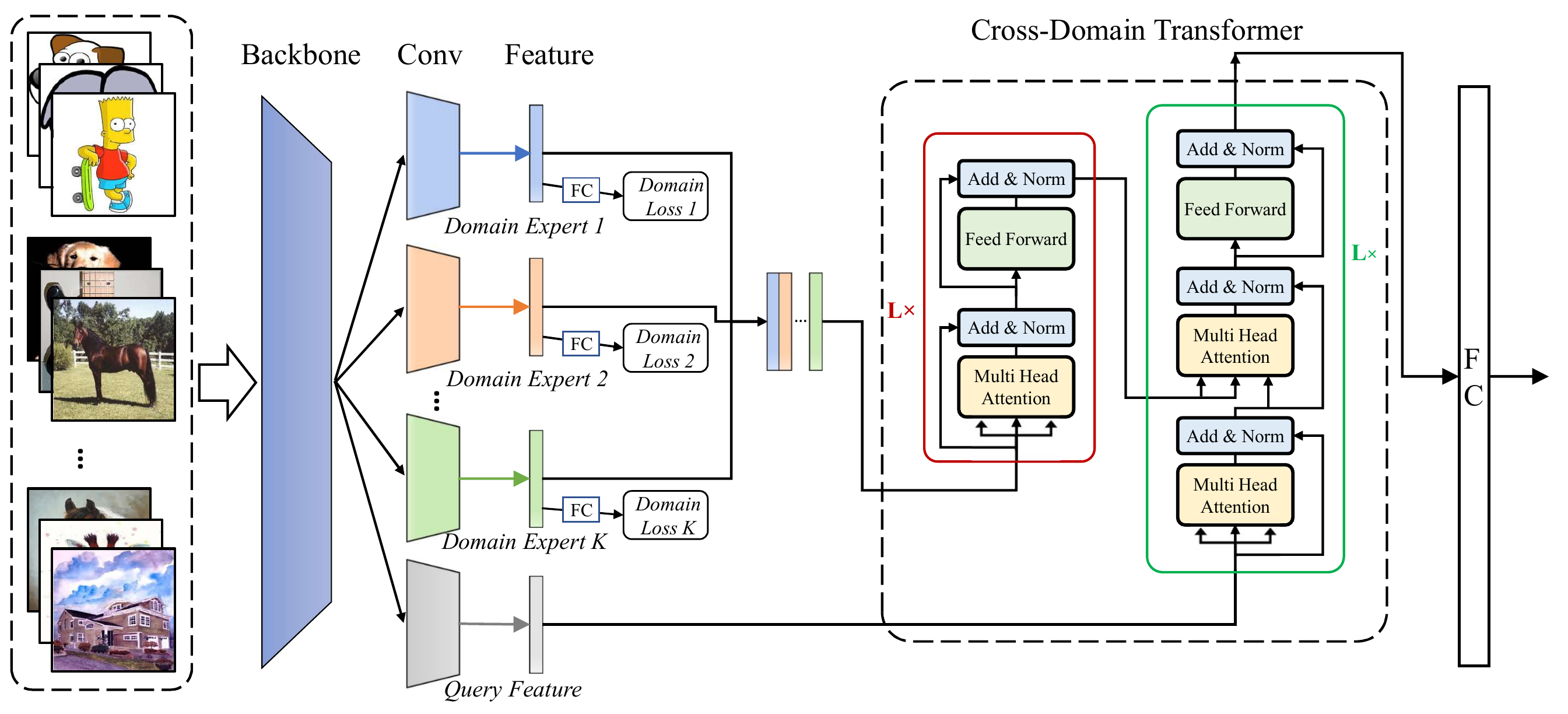}
\caption{The architecture of the proposed method D$^2$SDK.}
\label{fig:arch}
\end{figure*}

\subsection{The Proposed D$^2$SDK}

The overall architecture of the proposed method is shown in Figure \ref{fig:arch}. Each sample is passed through a shared backbone that acts as the common low-level feature extractor. The extracted low-level features are processed by $K$ local experts for domain-specific feature learning, each of which includes a feature extractor and a domain classifier. During training, each domain-specific loss is computed only based on samples belonging to that domain. Apart from the domain experts, a domain-agnostic feature extractor is also included to obtain the initial representation of the input sample. 
The domain-specific features are concatenated and fed into the Transformer encoder so that the Transformer is able to learn further dependency relationships and build the knowledge of existing source domains. The domain-agnostic features are directly presented to the Transformer decoder, which passes through a self-attention block before being fed into the cross-attention block as query features. At the same time, the outputs of Transformer encoder layers are presented as memory (i.e., keys and values) to the cross-attention block in the Transformer decoder layers. The relationships among different domains will be explored in the Transformer decoder layers, and the final learned representation is an ensemble of features across the source domains according to the dynamically learned domain relationships. A fully connected (FC) layer performs objection classification based on the decoded representation output by the cross-domain Transformer.

\noindent \textbf{Backbone}: The backbone is typically a CNN network (parameterized by $\vtheta^B$). The backbone is shared across all domain experts and the query feature branch. This enables common low-level CNN feature learning across different domains and avoids excessive computational and memory burden involved in building complete domain-specific networks. In our implementation, the ResNet architecture \cite{HeResNet2016} is utilized as the backbone due to its efficiency.

\noindent \textbf{Local Experts}: In order to learn discriminative features for each domain, we further append a convolutional neck layer at the beginning of each domain expert. The extracted low-level features are processed by these CNN necks to produce a $d$-dimensional domain-specific representation. A domain-specific classifier, which is a fully connected layer, is attached to each domain-specific CNN neck, which computes the domain-specific loss during training. The domain-specific feature extractor and classifier are together parameterized as $\vtheta^D_k$, $k = 1,2,\cdots,K$. A similar convolutional neck is also used as a domain-agnostic feature extractor, which outputs the query features for the cross-domain Transformer based on the input sample.

\noindent \textbf{Cross-Domain Transformer}: The $K$ domain-specific features from the local experts are fed to the Transformer encoder for source domain knowledge learning. Through the self-attention mechanism, the encoder is able to learn the dependency relationships and build knowledge about the available source domains. The output of the domain-agnostic feature branch is fed to the Transformer decoder as the query input to the cross-attention block. The output of the encoder is presented as memorized knowledge to the cross-attention block. The relationships among different domains will be explored in the Transformer decoder, and the resulting representation can be expected to encapsulate the discriminative features across all the source domains that are relevant for classification for the given input. This entire process can be considered as dynamic decoding of source domain knowledge (D$^2$SDK). The decoded representation is passed through a FC layer for classification. The domain-agnostic feature branch, cross-domain Transformer, and the final FC layer can be together considered as the global expert, whose parameters are denoted as $\vtheta^G$.

\subsection{Training and Inference}

The entire architecture is trained in an end-to-end fashion based on $\mathcal{D}_S$. The final loss function is a weighted combination of the domain expert loss and final classification (global expert) loss. The overall loss function for a training sample $(\vx_i,y_i,z_i)$ can be expressed as:
\begin{equation}\label{eqn:loss}
\begin{split}
   \mathcal{L}(\vx_i,y_i,z_i;\vtheta) =& \lambda\sum_{k=1}^{K}I(z_i=k)\mathcal{L}^D_k\left(g_k(\vx_i),y_i;\vtheta^{BD}_k\right) \\
  & + (1-\lambda)\mathcal{L}^G\left(h(\vx_i),y_i;\vtheta^{BG}\right),
\end{split}
\end{equation}
\noindent where $\vtheta = (\vtheta^B \cup \{\vtheta^D_k\}_{k=1}^{K} \cup \vtheta^G)$ is the set of all parameters in the proposed D$^2$SDK architecture, $\vtheta^{BG}$ represents all the parameters except those associated with the FC layers attached to the local experts, $\vtheta^{BD}_k = (\vtheta^B \cup \vtheta_k^D)$, $g_k(\vx_i)$ is the output of the domain-specific FC layer, $h(\vx_i)$ is the output of the final FC layer, and $I(.)$ is the indicator function, which takes value 1 if $z_i = k$ and 0 otherwise. Here, both the final classification loss $\mathcal{L}^G$ and the local domain expert loss $\mathcal{L}^D_k$ for domain $k$ are standard cross-entropy loss functions used for general classification and $\lambda$ ($0 \leq \lambda \leq 1$) is the relative weight assigned to the domain-specific loss.

During inference, the domain-specific classification heads are discarded. The given test image is passed through the rest of the D$^2$SDK framework and only the output of the final FC layer is used for classification. The interaction between the domain-specific features and the domain-agnostic feature within the cross-domain Transformer facilitates dynamic extraction of useful features, which provide good generalization performance on target domain images as demonstrated by the empirical results.

\section{Experiments}
\label{sec:experiments}

The domain generalization performance of the proposed D$^2$SDK framework is evaluated on three standard datasets and benchmarked against several state-of-the-art methods.

\subsection{Implementation Details}

In the experiments, ResNet \citep{HeResNet2016} until layer3 is used as the shared backbone. Instance normalization is used in the CNN backbone. Layer4 of ResNet is utilized as the convolutional neck layer in the domain-specific and domain-agnostic branches. In the Transformer, two layers of encoders and two layers of decoders are stacked ($L = 2$). The number of attention heads in the MHA block is set to two, and the features forwarded by the MLP block have a dimensionality of 1024. Sensitivity of the proposed method to Transformer parameters has been studied in section \ref{subsec:abla}. For the combined loss function in equation \ref{eqn:loss}, $\lambda$ is set to $0.1$ on all the evaluated datasets. For optimization, a stochastic gradient descent solver with a learning rate of $0.001$ and a batch size of $32$ is used. The training takes $80$ epochs. Following \cite{CarlucciJiGen2019}, the learning rate is decayed by a factor of $0.1$ after reaching $80$\% of the training epochs. The data augmentation strategy proposed in \citep{CarlucciJiGen2019} is also utilized on all evaluated datasets. The source code of this research will be publicly released. Finally, all reported performances have been averaged over ten rounds to avoid any random bias.

\begin{table}[t]
\centering
\begin{tabular}{l@{} | l l@{~} l@{~~} l l} \toprule
Method                                  & Photo & Artpainting & Cartoon & Sketch & Ave. \\ \midrule
    DSAM \cite{DInnocenteDSAM2018}      & 95.30 & 77.33 & 75.89 & 69.27 & 80.72 \\
    MLDG \cite{LiMLDG2018}              & 94.00 & 78.70 & 73.30 & 65.10 & 80.70 \\
    MetaReg \cite{BalajiMetaReg2018}    & 95.50 & 83.70 & 77.20 & 70.30 & 81.70 \\
    JiGen \cite{CarlucciJiGen2019}      & 96.03 & 79.42 & 75.25 & 71.35 & 80.51 \\
    MASF \cite{DouMASF2019}             & 94.99 & 80.29 & 71.17 & 71.69 & 81.03 \\
    AGG \cite{LiAGGEpiFCR2019}          & 94.40 & 77.60 & 73.90 & 70.30 & 79.10 \\
    Epifcr \cite{LiAGGEpiFCR2019}       & 93.90 & 82.10 & 77.00 & 73.00 & 81.50  \\
    CuMix \cite{ManciniCuMix2020}       & 95.10 & 82.30 & 76.50 & 72.60 & 81.60 \\
\small{MMLD \cite{MatsuuraMMLD2020}}     & 96.09 & 81.28 & 77.16 & 72.29 & 81.83 \\
    ER \cite{ZhaoER2020}                 & \textbf{96.65}& 80.70 & 76.40 & 71.77 & 81.38 \\
    DADG \cite{ChenDADG2022}            & 94.86 & 79.89 & 76.25 & 70.51 & 80.38 \\
\small{Mixstyle \cite{ZhouMixStyle2021}} & 96.10 & 84.10 & 78.80 & 75.90  & 83.70  \\
\small{L2A-OT \cite{ZhouLTAOT2020}}      & 96.20 & 83.30 & 78.20 & 73.60 & 82.80 \\
    DAEL \cite{ZhouDAEL2021}            & 95.60 & \textbf{84.60} & 74.40 & \textbf{78.90} & 83.40 \\ \midrule
    D$^2$SDK                            & 95.94 & 84.39 & \textbf{81.43} & 76.61 & \textbf{84.59} \\ \bottomrule

\end{tabular}
\caption{DG performance on PACS with ResNet-18 as backbone.}
\label{tlb:pacs}
\end{table}

\subsection{Datasets}

The proposed framework has been evaluated on PACS, Office-Home and VLCS datasets, which are the standard datasets used in domain generalization. The PACS dataset\footnote{\url{https://domaingeneralization.github.io/\#data}} \citep{LiArtierDG2017} contains 9,991 images coming from seven categories. It is composed of four domains, namely Photo (P), Art Painting (A), Cartoon (C) and Sketches (S). The experimental protocol proposed in \citep{LiArtierDG2017} for this benchmark is followed to ensure fair comparison.

The Office-Home dataset\footnote{\url{https://www.hemanthdv.org/officeHomeDataset.html}} \citep{VenkateswaraUDA2017} contains around 15,500 images from $N_C = 65$  categories comprising of daily use objects. It also has four domains, which are Art (A), Clipart (C), Product (P), and Real-World (R). The same protocol introduced for PACS is also followed for this dataset.

The VLCS dataset \citep{TorralbaUnbiasedLookAtDatasetBias2011} contains 10,729 images from five object categories that are shared by PASCAL VOC 2007 (V), Labelme (L), CALTECH (C) and SUN (S) databases which act as different domains. The experimental protocol proposed by \cite{GhifaryMTAE2015} is followed for a fair comparison.

Since all the three datasets have four domains and evaluation is based on the leave-one-domain-out DG experimental protocol, the number of source domains $K=3$ in all the experiments.

\begin{table}[t]
\centering
\begin{tabular}{ l@{} | l l l@{~~} l l} \toprule
Methods & Art   &   \small{Clipart} &  \small{Product} & \small{RWorld}   & Ave. \\ \midrule
    DSAM \cite{DInnocenteDSAM2018}   & 58.03 & 44.37 & 69.22 & 71.45 & 60.77 \\
    MLDG \cite{LiMLDG2018}           & 52.88 & 45.72 & 69.90 & 72.68 & 60.30 \\
    JiGen \cite{CarlucciJiGen2019}   & 53.04 & 47.51 & 71.47 & 72.79 & 61.20 \\
    DADG \cite{ChenDADG2022}         & 55.57 & 48.71 & 70.90 & 73.70 & 62.22 \\
\small{L2A-OT \cite{ZhouLTAOT2020}}  & \textbf{60.60} & 50.10 & 74.80 & 77.00 & 65.60 \\
    DAEL \cite{ZhouDAEL2021}         & 59.40 & \textbf{55.10} & 74.00 & 75.70 & 66.10 \\ \midrule
    D$^2$SDK                         & 60.34 & 52.23 & \textbf{75.05}& \textbf{77.64} & \textbf{66.32} \\ \bottomrule
\end{tabular}
\caption{DG performance on Office-Home with ResNet-18 as backbone.}
\label{tlb:office}
\end{table}

\begin{table}[t]
\centering
\begin{tabular}{ l@{}| l l@{~~} l l l }
\toprule
Method                                      & \small{Caltech} & \small{Labelme}  & \small{Pascal}    & Sun   &  Ave.   \\   \midrule
\footnotesize{MMDAAE\cite{LiMMD2018}} & 94.40       & 62.60       & 67.70       & 64.40       & 72.30      \\
D-SAM \cite{DInnocenteDSAM2018}             & 91.75       & 56.95       & 58.59       & 60.84       & 67.03       \\
MLDG \cite{LiMLDG2018}                      & 94.40       & 61.30       & 67.70       & 65.90       & 72.30      \\
JiGen \cite{CarlucciJiGen2019}              & 96.93       & 60.90       & 70.62       & 64.30       & 73.19       \\
AGG \cite{LiAGGEpiFCR2019}                  & 93.10       & 60.60       & 65.40       & 65.80       & 71.20       \\
Epifcr \cite{LiAGGEpiFCR2019}               & 94.10       & 64.30       & 67.10       & 65.90       & 72.90       \\
MASF \cite{DouMASF2019}                     & 94.78       &   64.90     &   69.14     &   67.64     &   74.11     \\
DADG \cite{ChenDADG2022}                    & 96.80 & 66.81 & 70.77 & 63.64 & 74.46 \\ \midrule
D$^2$SDK & 97.41 & 62.63 & 75.48 & 69.04 & 76.14 \\ \bottomrule
\end{tabular}
\caption{DG performance on VLCS.}
\label{tlb:vlcs}
\end{table}

\subsection{Empirical Results}

\textbf{PACS}: DG performance of the proposed method on the PACS dataset with ResNet-18 as the backbone is shown in Table \ref{tlb:pacs}. The proposed  D$^2$SDK method achieves best average performance compared to other benchmarked methods as well as the best performance on the Cartoon target domain. It achieves a clear improvement compared to the second best performance by DAEL \citep{ZhouDAEL2021}.

\textbf{Office-Home}: Results on the Office-Home dataset with ResNet-18 as the backbone are shown in Table \ref{tlb:office}. It can be observed that the D$^2$SDK method outperforms the state-of-the-art methods (LTA-OT \citep{ZhouLTAOT2020} and DAEL \citep{ZhouDAEL2021}) on average and also achieves the best performance on two of the four target domains.

\textbf{VLCS}: The results on the VLCS dataset are shown in Table \ref{tlb:vlcs}. While D$^2$SDK achieves the best performance, the comparison may not be very fair because the proposed method uses a ResNet-18 backbone and the other benchmarked methods in Table \ref{tlb:vlcs} use an AlexNet backbone.

The experimental results on the three datasets demonstrate that the proposed method has a good domain generalization ability for unseen domains, thanks to the cross-domain Transformer mechanism, where the source domain knowledge is encoded and dynamically decoded for the inference of new images from unseen domains.

\begin{table}[t]
\centering
\begin{tabular}{  l@{~}| l@{~~} l@{~} l@{~~} l@{~~} l  } \toprule
\multicolumn{6}{c}{\emph{PACS Dataset}} \\ \midrule
Methods     & Photo & \small{Artpainting} & Cartoon  & Sketch & Ave. \\ \hline
ResNet18              & 95.19  & 80.41 & 75.94 & 73.28 & 81.20  \\
D$^2$SDK$_{18}$       & 95.94 & 84.39 & 81.43 & 76.61 & 84.59 \\
ResNet50              & 97.09  & 87.29 & 81.00 & 74.11 & 84.87 \\
D$^2$SDK$_{50}$       & 97.47  & 88.67 & 84.96 & 78.78 & 87.47 \\ \midrule
\multicolumn{6}{c}{\emph{Office-Home Dataset}}\\ \midrule
Methods               & Art   &   Clipart &  Product & RWorld   & Ave. \\ \hline
ResNet18              & 54.06  & 47.56 & 72.17& 74.22 & 62.00   \\
D$^2$SDK$_{18}$       & 60.34  & 52.23 & 75.05 & 77.64  & 66.32 \\
ResNet50              & 62.09 & 53.29 & 76.67 & 78.67 & 67.68 \\
D$^2$SDK$_{50}$    & 68.92 & 57.62 & 80.25 & 82.27 & 72.26   \\ \midrule
\multicolumn{6}{c}{\emph{VLCS Dataset}}\\ \midrule
Methods     & Caltech     & Labelme     &   Pascal    & Sun   &  Ave.   \\ \hline
ResNet18          & 96.55 & 62.50 & 72.19 & 66.52 & 74.44     \\
D$^2$SDK$_{18}$    & 97.41 & 62.63 & 75.48 & 69.04 & 76.14  \\
ResNet50          & 98.34 & 63.13 & 74.50 & 69.93  & 76.47  \\
D$^2$SDK$_{50}$    & 97.47 & 63.03 & 78.12 & 70.53 & 77.29 \\ \bottomrule
\end{tabular}
\caption{Experimental results with comparison to baselines. D$^2$SDK$_{18}$ is with ResNet-18 while D$^2$SDK$_{50}$ is with ResNet-50.}
\label{tlb:all50}
\end{table}

Note that the state-of-the-art performance on the PACS dataset (based on ResNet-18 backbone) reported in the literature is marginally (less than $1$\%) higher than that of the proposed D$^2$SDK method. The DSON \citep{SeoDSN2020} and RSC \citep{HuangSelfChallenging2020} methods report an average accuracy of $85.11$\% and $85.15$\%, respectively. However, the accuracies of these two methods on the Office-Home dataset are only $62.90$\% (DSON) and $63.12$\% (RSC), which are more than $3$\% lower than the accuracy of the D$^2$SDK method. Compared to PACS and VLCS datasets that have small number of object categories ($7$ and $5$, respectively) and are easily prone to local minima, the Office-Home dataset with $65$ object categories is a more challenging dataset and may be more reliable for evaluation of domain generalization algorithms. Thus, the clear superiority of the D$^2$SDK method on this dataset strongly demonstrates the benefits of the proposed approach. It must also be emphasized that the proposed method performs better than other state-of-the-art MoE algorithms for DG such as D-SAM \citep{DInnocenteDSAM2018}, DSON \citep{SeoDSN2020}, and DAEL \citep{ZhouDAEL2021}.

Furthermore, our analysis of the published code indicates that the RSC method \citep{HuangSelfChallenging2020} stops training early based on the best epoch performance on target domain. In contrast, all the results reported for the proposed method are based on the model learned at the last epoch in training (with strictly no access to target domain data). A detailed analysis of this model selection issue has been presented in section \ref{subsec:more}.

\subsection{Ablation Studies}
\label{subsec:abla}

\noindent \textbf{Impact of Cross-Domain Transformer module}: In Table \ref{tlb:all50}, the proposed D$^2$SDK method is compared against baseline ResNet-18 and ResNet-50 architectures. Since the baseline architectures are directly trained on the available source domains in a domain-agnostic manner, the baseline results closely approximate the performance of a classifier that works only on the domain-agnostic (query) features. It can be observed that D$^2$SDK performs better than the baselines by more than $4$\% on average, especially on PACS and Office-Home datasets. This improvement is solely due to the inclusion of the cross-domain Transformer, which once again demonstrates the ability of this module to extract more robust features that generalize well across domains. Moreover, since the proposed approach is not tied to the baseline architecture, it can be used in conjunction with better baselines. For example, changing the backbone from ResNet-18 to ResNet-50 or training the baseline using more sophisticated methods (e.g., \cite{ChaSWAD2021}) can lead to further performance improvement.

\begin{table}[t]
\centering
\begin{tabular}{ l@{}| l l@{~} l@{~} l l } \toprule
Model & \small{Photo} & \small{Artpainting}  & \small{Cartoon}  & \small{Sketch} & Ave. \\\midrule
ConvExp         & 95.35 & 81.23 & 76.13 & 73.26 & 81.49  \\
TEExp           & 94.69 & 81.99 & 76.42 & 73.96 & 81.77   \\
TD-D$^2$SDK	& 95.61 & 83.18 & 81.27 & 77.31 & 84.34   \\ \bottomrule
\end{tabular}
\caption{Performance of sub-models of the D$^2$SDK architecture.}
\label{tlb:submodel}
\end{table}

\noindent \textbf{Impact of Transformer Encoder and Decoder}: In order to evaluate the contributions of individual components within the proposed architecture, we study the performance of three sub-models on the PACS dataset. The first sub-model performs inference based on the sum of domain-specific outputs. Since it is a naive ensemble architecture with no Transformer, we denote it as ConvExp. In the second sub-model, the Transformer decoder is dropped (while retaining the shared backbone, domain experts, and Transformer encoder) and the outputs of the Transformer encoder are fed to separate domain-specific classifiers, whose outputs are summed for inference. This second sub-model is referred to as Transformer Encoder Expert (TEExp). Finally, in the third sub-model called TD-D$^2$SDK, the Transformer encoder is ignored (while retaining the shared backbone, domain experts, and Transformer decoder) and the domain-specific representations are directly fed as memory (keys and values) to the Transformer decoder. For experiments with all the three sub-models, the saved D$^2$SDK model with ResNet-18 backbone is used for initialization and the results are reported in Table \ref{tlb:submodel}.

\begin{table}[t]
\centering
\begin{tabular}{l | l l@{~} l l l} \toprule
$\lambda$ & Photo & \small{Artpainting}  & Cartoon  & Sketch & Ave. \\\midrule
0.7	   & 93.35&	81.30&	81.40&	79.98&	84.00 \\
0.5	   & 94.48&	82.80&	81.65&	78.51&	84.36 \\
0.3	   & 95.32&	83.71&	81.43&	77.34&	84.45 \\
0.2	   & 95.65&	84.12&	80.86&	78.36&	84.74 \\
0.1	   & 95.94& 84.39&  81.43&  76.61&  84.59 \\
0.05   & 95.70&	83.64&	80.87&	76.91&	84.28 \\
0.02   & 95.64&	83.75&	81.43&	77.35&	84.54 \\
0.01   & 95.71&	84.05&	81.21&	76.40&	84.34 \\ \bottomrule
\end{tabular}
\caption{Sensitivity of proposed method to parameter $\lambda$.}
\label{tlb:lambda}
\end{table}

From Table \ref{tlb:submodel}, it can be observed that the Transformer encoder with domain experts (TEExp) sub-model has a marginally better performance compared to the ConvExp model. In contrast, the TD-D$^2$SDK sub-model (which skips the Transformer encoder) has comparable performance to the full D$^2$SDK method. This shows that the dynamic decoding procedure based on cross-attention mechanism plays the most critical role in the proposed architecture. Thus, it may be possible to simplify the proposed architecture (by removing the Transformer encoder) to reduce computations and achieve comparable performance in practice.

\noindent \textbf{Sensitivity to parameter} $\lambda$: The influence of parameter $\lambda$ (the weight assigned to the domain expert loss) is studied on the PACS dataset with ResNet-18 as the backbone. The results in Table \ref{tlb:lambda} indicate that except when $\lambda$ is too large ($> 0.5$), the overall average accuracy is quite stable, indicating that the D$^2$SDK method is relatively insensitive to this parameter. Therefore, $\lambda$ is simply set to $0.1$ for all the experiments.

\noindent \textbf{Sensitivity to Transformers parameters}: The impact of Transformer parameters in the proposed architecture is also studied on PACS dataset with ResNet-18 as backbone. We consider three parameters: the Transformer depth ($L$, the number of layers in the encoder and decoder stacks), the number of attention heads in the MHA block, and the feed forward feature dimension in the MLP block. For convenience, the pre-trained backbone is used in the study. Experimental results shown in Table \ref{tlb:param} indicate that all the three Transformer parameters have negligible impact on DG performance on PACS dataset. This could be because Transformers typically require very large number of samples to show their learning ability, while PACS is a small dataset. Moreover, since the proposed architecture uses a shared CNN network as backbone, a large Transformer may not be required for this DG task.

\begin{table}[t]
\centering
\begin{tabular}{l@{} | l l@{~} l@{~~} l l} \toprule
Params & Photo & \small{Artpainting}  & Cartoon  & Sketch & Ave. \\\midrule
L = 2      & 94.94 & 80.94&	78.54&	72.48&	81.73   \\
L = 3      & 94.81 & 81.67&	78.88&	72.87&	82.06   \\
L = 4      & 94.74 & 81.23&	79.18&	72.81&	81.99   \\
L = 5	   & 94.49 & 80.43&	78.17&	72.78&	81.47   \\ \midrule
Head2	   & 94.94&	80.94&	78.54&	72.48&	81.73  \\
Head4      & 94.62&	81.33&	79.29&	72.84&	82.02   \\
Head8      & 94.81&	82.21&	78.60&	71.04&	81.66   \\
Head16     & 94.86&	81.78&	78.41&	73.10&	82.04   \\ \midrule
512-D	   & 94.77	&80.79	&78.19	&73.51	&81.82  \\
1024D      & 94.94	&80.94	&78.54	&72.48	&81.73 \\
2048D      & 94.68	&81.30	&78.29	&72.52	&81.70  \\
4096D      & 94.73	&81.66	&78.45	&73.22	&82.02  \\\bottomrule
\end{tabular}
\caption{DG performance with different Transformer parameter settings.}
\label{tlb:param}
\end{table}

\noindent \textbf{Limitations}: One of the main disadvantages of the proposed method is the higher computational complexity of the inference process in comparison to the baseline architecture. However, this is a common problem for most MoE methods, because multiple branches are required and a subnetwork is required for predicting aggregation weights. Furthermore, ablation study of the proposed method shows a possible way to reduce the computational cost by discarding the Transformer encoder. It may also be possible to eliminate the self-attention block at the beginning of the Transformer decoder. However, further studies may be required to verify these observations more carefully.

\subsection{Model Selection Results}
\label{subsec:more}


Most DG experimental protocols are based on leave-one-domain-out model selection \citep{GulrajaniISLDG2021}, where no information from the target domain is provided during training. The source domain images are divided into training and validation sets, and the learned model is tested on the unseen target domain as test set. In practice, we found that the best model selected on the validation set and the finally learned model at the last epoch of training have a similar performance on the target domain. However, these results are not usually as good as the best performance among all epochs monitored on the test set. We refer to this procedure as test-set best-epoch performance. To the best of our knowledge, some existing methods \citep{HuangSelfChallenging2020} report this kind of best-epoch results monitored on the test set, which is not quite fair. In the main text of this work, the results reported for the proposed method are based on the learned model at the last epoch. In this subsection, we would like to additionally provide the test-set best-epoch performance as well (see Table \ref{tlb:best}), in case there is a need for such kind of comparison. However, we strongly discourage doing so because such oracle model selection \citep{GulrajaniISLDG2021} criteria may lead to overly optimistic performance results. Comparing Table \ref{tlb:best} and Table \ref{tlb:all50}, it can be observed that the test-set best-epoch results on PACS and VLCS are clearly better than our last-epoch results, with about $2$\%-$3$\% performance gap. For the Office-Home dataset, the performance of the finally learned model is close to the best-epoch results in Table \ref{tlb:best}. As explained earlier, PACS contains only seven categories and VLCS contains only five categories, so there are more chances to gain better performance due to local minima. 

\begin{table}[t]
\centering
\begin{tabular}{  l@{~}| l@{~~} l@{~} l@{~~} l@{~~} l  } \toprule
\multicolumn{6}{c}{\emph{PACS Dataset}} \\ \midrule
Methods     & Photo & \small{Artpaiting} & Cartoon & Sketch & Ave. \\ \hline
D$^2$SDK$_{18}$ & 96.69 & 85.99 & 84.30 & 79.90 & 86.72  \\
D$^2$SDK$_{50}$ & 98.07 & 90.64 & 86.75 & 81.76 & 89.30   \\ \midrule
\multicolumn{6}{c}{\emph{Office-Home Dataset}}\\ \midrule
Methods     & Art   &   Clipart &  Product & RWorld   & Ave. \\ \hline
D$^2$SDK$_{18}$ & 61.01 & 53.20 & 75.44 & 77.81 & 66.86    \\
D$^2$SDK$_{50}$ & 69.53 & 59.11 & 80.55 & 82.56 & 72.94  \\ \midrule
\multicolumn{6}{c}{\emph{VLCS Dataset}}\\ \midrule
Methods     & Caltech     & Labelme     &   Pascal    & Sun   &  Ave.   \\ \hline
D$^2$SDK$_{18}$ & 98.94 & 67.52 & 78.77 & 72.09 & 79.33  \\
D$^2$SDK$_{50}$ & 99.43 & 67.65 & 81.36 & 74.93 & 80.84  \\ \bottomrule
\end{tabular}
\caption{The test-set best-epoch performance on three datasets.}
\label{tlb:best}
\end{table}

\section{Conclusions}\label{sec:conclusions}
This paper shows that given domain-specific local experts and domain-agnostic query features of the test sample, Transformers are effective in discovering domain relationships and in turn help with accurate inference on images from unseen domains. This is possible thanks to the attention mechanism in the Transformers. In the proposed method, while the encoder learns the source domain knowledge and builds the memory, the decoder dynamically decodes the learned source domain knowledge to extract better features for the given test sample from the unseen target domain. Thus, the proposed approach represents a promising research direction towards solving the domain generalization challenge. Future work will focus on making suitable modifications to the Transformer architecture to reduce its complexity, while simultaneously enhancing its ability to model more diverse and fine-grained domain shifts.


\bibliographystyle{iclr2022_conference}
\bibliography{dgbib}

\begin{thebibliography}{39}
\providecommand{\natexlab}[1]{#1}
\providecommand{\url}[1]{\texttt{#1}}
\expandafter\ifx\csname urlstyle\endcsname\relax
  \providecommand{\doi}[1]{doi: #1}\else
  \providecommand{\doi}{doi: \begingroup \urlstyle{rm}\Url}\fi

\bibitem[Balaji et~al.(2018)Balaji, Sankaranarayanan, and
  Chellappa]{BalajiMetaReg2018}
Yogesh Balaji, Swami Sankaranarayanan, and Rama Chellappa.
\newblock {MetaReg: Towards Domain Generalization using Meta-Regularization}.
\newblock In \emph{{NeurIPS}}, volume~31, 2018.

\bibitem[Carion et~al.(2020)Carion, Massa, Synnaeve, Usunier, Kirillov, and
  Zagoruyko]{CarionDETR2020}
Nicolas Carion, Francisco Massa, Gabriel Synnaeve, Nicolas Usunier, Alexander
  Kirillov, and Sergey Zagoruyko.
\newblock {End-to-End Object Detection with Transformers}.
\newblock In \emph{{ECCV}}, pp.\  213--229, 2020.

\bibitem[Carlucci et~al.(2019)Carlucci, D'Innocente, Bucci, Caputo, and
  Tommasi]{CarlucciJiGen2019}
Fabio~M. Carlucci, Antonio D'Innocente, Silvia Bucci, Barbara Caputo, and
  Tatiana Tommasi.
\newblock {Domain Generalization by Solving Jigsaw Puzzles}.
\newblock In \emph{{CVPR}}, 2019.

\bibitem[Cha et~al.(2021)Cha, Chun, Lee, Cho, Park, Lee, and Park]{ChaSWAD2021}
Junbum Cha, Sanghyuk Chun, Kyungjae Lee, Han-Cheol Cho, Seunghyun Park, Yunsung
  Lee, and Sungrae Park.
\newblock {SWAD: Domain Generalization by Seeking Flat Minima}.
\newblock In \emph{{NeurIPS}}, 2021.

\bibitem[Chen et~al.(2022)Chen, Zhuang, and Chang]{ChenDADG2022}
Keyu Chen, Di~Zhuang, and J.~Morris Chang.
\newblock {Discriminative adversarial domain generalization with meta-learning
  based cross-domain validation}.
\newblock \emph{Neurocomputing}, 467:\penalty0 418--426, 2022.

\bibitem[Dai et~al.(2021)Dai, Li, Liu, Tong, and Duan]{DaiRAMoE2021}
Yongxing Dai, Xiaotong Li, Jun Liu, Zekun Tong, and Ling-Yu Duan.
\newblock {Generalizable Person Re-Identification With Relevance-Aware Mixture
  of Experts}.
\newblock In \emph{{CVPR}}, pp.\  16145--16154, 2021.

\bibitem[D’Innocente \& Caputo(2018)D’Innocente and
  Caputo]{DInnocenteDSAM2018}
Antonio D’Innocente and Barbara Caputo.
\newblock {Domain generalization with domain-specific aggregation modules}.
\newblock In \emph{{German Conference on Pattern Recognition}}, pp.\  187--198,
  2018.

\bibitem[Dosovitskiy et~al.(2021)Dosovitskiy, Beyer, Kolesnikov, Weissenborn,
  Zhai, Unterthiner, Dehghani, Minderer, Heigold, Gelly, Uszkoreit, and
  Houlsby]{DosovitskiyViT2021}
Alexey Dosovitskiy, Lucas Beyer, Alexander Kolesnikov, Dirk Weissenborn,
  Xiaohua Zhai, Thomas Unterthiner, Mostafa Dehghani, Matthias Minderer, Georg
  Heigold, Sylvain Gelly, Jakob Uszkoreit, and Neil Houlsby.
\newblock {An Image is Worth 16x16 Words: Transformers for Image Recognition at
  Scale}.
\newblock In \emph{ICLR}, 2021.

\bibitem[Dou et~al.(2019)Dou, Coelho~de Castro, Kamnitsas, and
  Glocker]{DouMASF2019}
Qi~Dou, Daniel Coelho~de Castro, Konstantinos Kamnitsas, and Ben Glocker.
\newblock {Domain Generalization via Model-Agnostic Learning of Semantic
  Features}.
\newblock In \emph{{NeurIPS}}, volume~32, 2019.

\bibitem[Finn et~al.(2017)Finn, Abbeel, and
  Levine]{FinnModelAgnosticMetaLearning2017}
Chelsea Finn, Pieter Abbeel, and Sergey Levine.
\newblock {Model-Agnostic Meta-Learning for Fast Adaptation of Deep Networks}.
\newblock In \emph{{ICML}}, pp.\  1126–1135, 2017.

\bibitem[Ganin et~al.(2016)Ganin, Ustinova, Ajakan, Germain, Larochelle,
  Laviolette, Marchand, and Lempitsky]{GaninDomainAdversarialTraining2016}
Yaroslav Ganin, Evgeniya Ustinova, Hana Ajakan, Pascal Germain, Hugo
  Larochelle, François Laviolette, Mario Marchand, and Victor Lempitsky.
\newblock {Domain-Adversarial Training of Neural Networks}.
\newblock \emph{{JMLR}}, 17, 2016.

\bibitem[Ghifary et~al.(2015)Ghifary, Kleijn, Zhang, and
  Balduzzi]{GhifaryMTAE2015}
Muhammad Ghifary, W~Bastiaan Kleijn, Mengjie Zhang, and David Balduzzi.
\newblock {Domain generalization for object recognition with multi-task
  autoencoders}.
\newblock In \emph{{ICCV}}, pp.\  2551--2559, 2015.

\bibitem[Gulrajani \& Lopez-Paz(2021)Gulrajani and
  Lopez-Paz]{GulrajaniISLDG2021}
Ishaan Gulrajani and David Lopez-Paz.
\newblock {In Search of Lost Domain Generalization}.
\newblock In \emph{{ICLR}}, 2021.

\bibitem[He et~al.(2016)He, Zhang, Ren, and Sun]{HeResNet2016}
Kaiming He, Xiangyu Zhang, Shaoqing Ren, and Jian Sun.
\newblock {Deep residual learning for image recognition}.
\newblock In \emph{{CVPR}}, pp.\  770--778, 2016.

\bibitem[Huang et~al.(2020)Huang, Wang, Xing, and
  Huang]{HuangSelfChallenging2020}
Zeyi Huang, Haohan Wang, Eric~P. Xing, and Dong Huang.
\newblock {Self-challenging improves cross-domain generalization}.
\newblock In \emph{{ECCV}}, pp.\  124--140, 2020.

\bibitem[Khosla et~al.(2012)Khosla, Zhou, Malisiewicz, Efros, and
  Torralba]{KhoslaUndoingDatasetBias2012}
Aditya Khosla, Tinghui Zhou, Tomasz Malisiewicz, Alexei~A. Efros, and Antonio
  Torralba.
\newblock {Undoing the Damage of Dataset Bias}.
\newblock In \emph{{ECCV}}, pp.\  158--171, 2012.

\bibitem[Li et~al.(2017)Li, Yang, Song, and Hospedales]{LiArtierDG2017}
Da~Li, Yongxin Yang, Yi-Zhe Song, and Timothy~M. Hospedales.
\newblock Deeper, broader and artier domain generalization.
\newblock In \emph{{ICCV}}, pp.\  5543--5551, 2017.

\bibitem[Li et~al.(2018{\natexlab{a}})Li, Yang, Song, and
  Hospedales]{LiMLDG2018}
Da~Li, Yongxin Yang, Yi-Zhe Song, and Timothy~M. Hospedales.
\newblock {Learning to generalize: Meta-learning for domain generalization}.
\newblock In \emph{{AAAI}}, 2018{\natexlab{a}}.

\bibitem[Li et~al.(2019)Li, Zhang, Yang, Liu, Song, and
  Hospedales]{LiAGGEpiFCR2019}
Da~Li, Jianshu Zhang, Yongxin Yang, Cong Liu, Yi-Zhe Song, and Timothy~M.
  Hospedales.
\newblock {Episodic training for domain generalization}.
\newblock In \emph{{ICCV}}, pp.\  1446--1455, 2019.

\bibitem[Li et~al.(2018{\natexlab{b}})Li, Pan, Wang, and Kot]{LiMMD2018}
Haoliang Li, Sinno~Jialin Pan, Shiqi Wang, and Alex~C. Kot.
\newblock {Domain Generalization with Adversarial Feature Learning}.
\newblock In \emph{{CVPR}}, pp.\  5400--5409, 2018{\natexlab{b}}.

\bibitem[Mancini et~al.(2018)Mancini, Rota~Bul\`o, Caputo, and
  Ricci]{ManciniBestSourcesForward2018}
Massimilano Mancini, Samuel Rota~Bul\`o, Barbara Caputo, and Elisa Ricci.
\newblock {Best sources forward: domain generalization through source-specific
  nets}.
\newblock In \emph{{ICIP}}, 2018.

\bibitem[Mancini et~al.(2020)Mancini, Akata, Ricci, and
  Caputo]{ManciniCuMix2020}
Massimiliano Mancini, Zeynep Akata, Elisa Ricci, and Barbara Caputo.
\newblock {Towards Recognizing Unseen Categories in Unseen Domains}.
\newblock In \emph{{ECCV}}, 2020.

\bibitem[Matsuura \& Harada(2020)Matsuura and Harada]{MatsuuraMMLD2020}
Toshihiko Matsuura and Tatsuya Harada.
\newblock {Domain Generalization Using a Mixture of Multiple Latent Domains}.
\newblock In \emph{{AAAI}}, pp.\  11749--11756, 2020.

\bibitem[Motiian et~al.(2017)Motiian, Piccirilli, Adjeroh, and
  Doretto]{MotiianDeepSDA2017}
Saeid Motiian, Marco Piccirilli, Donald~A. Adjeroh, and Gianfranco Doretto.
\newblock {Unified deep supervised domain adaptation and generalization}.
\newblock In \emph{{ICCV}}, pp.\  5715--5725, 2017.

\bibitem[Muandet et~al.(2013)Muandet, Balduzzi, and
  Sch\"{o}lkopf]{MuandetInvariantFeatures2013}
Krikamol Muandet, David Balduzzi, and Bernhard Sch\"{o}lkopf.
\newblock {Domain Generalization via Invariant Feature Representation}.
\newblock In \emph{{ICML}}, 2013.

\bibitem[Santoro et~al.(2016)Santoro, Bartunov, Botvinick, Wierstra, and
  Lillicrap]{SantoroMetaLearningMANN2016}
Adam Santoro, Sergey Bartunov, Matthew Botvinick, Daan Wierstra, and Timothy
  Lillicrap.
\newblock {Meta-Learning with Memory-Augmented Neural Networks}.
\newblock In \emph{{ICML}}, pp.\  1842–1850, 2016.

\bibitem[Seo et~al.(2020)Seo, Suh, Kim, Kim, Han, and Han]{SeoDSN2020}
Seonguk Seo, Yumin Suh, Dongwan Kim, Geeho Kim, Jongwoo Han, and Bohyung Han.
\newblock {Learning to Optimize Domain Specific Normalization for Domain
  Generalization}.
\newblock In \emph{{ECCV}}, pp.\  68--83, 2020.

\bibitem[Shankar et~al.(2018)Shankar, Piratla, Chakrabarti, Chaudhuri, Jyothi,
  and Sarawagi]{ShankarCrossGrad2018}
Shiv Shankar, Vihari Piratla, Soumen Chakrabarti, Siddhartha Chaudhuri, Preethi
  Jyothi, and Sunita Sarawagi.
\newblock {Generalizing across domains via cross-gradient training}.
\newblock In \emph{{ICLR}}, 2018.

\bibitem[Srinivas et~al.(2021)Srinivas, Lin, Parmar, Shlens, Abbeel, and
  Vaswani]{SrinivasBottleneckTransformer2021}
Aravind Srinivas, Tsung-Yi Lin, Niki Parmar, Jonathon Shlens, Pieter Abbeel,
  and Ashish Vaswani.
\newblock {Bottleneck transformers for visual recognition}.
\newblock In \emph{{CVPR}}, pp.\  16519--16529, 2021.

\bibitem[Torralba \& Efros(2011)Torralba and
  Efros]{TorralbaUnbiasedLookAtDatasetBias2011}
Antonio Torralba and Alexei~A. Efros.
\newblock {Unbiased look at dataset bias}.
\newblock In \emph{{CVPR}}, pp.\  1521--1528, 2011.

\bibitem[Vaswani et~al.(2017)Vaswani, Shazeer, Parmar, Uszkoreit, Jones, Gomez,
  Kaiser, and Polosukhin]{VaswaniTransformers2017}
Ashish Vaswani, Noam Shazeer, Niki Parmar, Jakob Uszkoreit, Llion Jones,
  Aidan~N Gomez, {\L}ukasz Kaiser, and Illia Polosukhin.
\newblock {Attention is all you need}.
\newblock In \emph{{NeurIPS}}, pp.\  5998--6008, 2017.

\bibitem[Venkateswara et~al.(2017)Venkateswara, Eusebio, Chakraborty, and
  Panchanathan]{VenkateswaraUDA2017}
Hemanth Venkateswara, Jose Eusebio, Shayok Chakraborty, and Sethuraman
  Panchanathan.
\newblock {Deep hashing network for unsupervised domain adaptation}.
\newblock In \emph{{CVPR}}, pp.\  5018--5027, 2017.

\bibitem[Volpi et~al.(2018)Volpi, Namkoong, Sener, Duchi, Murino, and
  Savarese]{VolpiADA2018}
Riccardo Volpi, Hongseok Namkoong, Ozan Sener, John Duchi, Vittorio Murino, and
  Silvio Savarese.
\newblock {Generalizing to unseen domains via adversarial data augmentation}.
\newblock In \emph{{NeurIPS}}, 2018.

\bibitem[Wang et~al.(2021)Wang, Lan, Liu, Ouyang, Zeng, and
  Qin]{WangDGSurvey2021}
Jindong Wang, Cuiling Lan, Chang Liu, Yidong Ouyang, Wenjun Zeng, and Tao Qin.
\newblock {Generalizing to Unseen Domains: A Survey on Domain Generalization}.
\newblock In \emph{{IJCAI}}, 2021.

\bibitem[Zhao et~al.(2020)Zhao, Gong, Liu, Fu, and Tao]{ZhaoER2020}
Shanshan Zhao, Mingming Gong, Tongliang Liu, Huan Fu, and Dacheng Tao.
\newblock {Domain generalization via entropy regularization}.
\newblock In \emph{{NeurIPS}}, volume~33, 2020.

\bibitem[Zhou et~al.(2020)Zhou, Yang, Hospedales, and Xiang]{ZhouLTAOT2020}
Kaiyang Zhou, Yongxin Yang, Timothy~M. Hospedales, and Tao Xiang.
\newblock {Learning to Generate Novel Domains for Domain Generalization}.
\newblock In \emph{{ECCV}}, pp.\  561--578, 2020.

\bibitem[Zhou et~al.(2021{\natexlab{a}})Zhou, Liu, Qiao, Xiang, and
  Loy]{ZhouDGSurveyVision2021}
Kaiyang Zhou, Ziwei Liu, Yu~Qiao, Tao. Xiang, and Chen~C. Loy.
\newblock {Domain Generalization in Vision: A Survey}, 2021{\natexlab{a}}.

\bibitem[Zhou et~al.(2021{\natexlab{b}})Zhou, Yang, Qiao, and
  Xiang]{ZhouDAEL2021}
Kaiyang Zhou, Yongxin Yang, Yu~Qiao, and Tao Xiang.
\newblock {Domain Adaptive Ensemble Learning}.
\newblock \emph{IEEE Transactions on Image Processing}, 30:\penalty0
  8008--8018, 2021{\natexlab{b}}.

\bibitem[Zhou et~al.(2021{\natexlab{c}})Zhou, Yang, Qiao, and
  Xiang]{ZhouMixStyle2021}
Kaiyang Zhou, Yongxin Yang, Yu~Qiao, and Tao Xiang.
\newblock {Domain Generalization With MixStyle}.
\newblock In \emph{{ICLR}}, 2021{\natexlab{c}}.

\end{thebibliography}

\end{document}